\pdfoutput=1
\PassOptionsToPackage{dvipsnames}{xcolor}
\documentclass[11pt]{article}
\usepackage{acl}
\usepackage{mathtools}
\usepackage{times}
\usepackage{latexsym}

\usepackage{amsmath,amsfonts,bm}

\def\eqref#1{equation~\ref{#1}}

\def\1{\bm{1}}

\def\vh{{\bm{h}}}

\DeclareMathAlphabet{\mathsfit}{\encodingdefault}{\sfdefault}{m}{sl}
\SetMathAlphabet{\mathsfit}{bold}{\encodingdefault}{\sfdefault}{bx}{n}

 \usepackage[T1]{fontenc}

\usepackage[utf8]{inputenc}
\usepackage{xspace}
\usepackage{booktabs,amsfonts,dcolumn}
\newcommand\tf[1]{\textbf{#1}}

\newcommand{\ba}{$_\texttt{base}$}
\newcommand{\cls}{{\tt[CLS]}}
\newcommand{\ours}{SCD\xspace}
\usepackage{microtype}
\usepackage{amsmath}
\usepackage{arydshln}
\usepackage{multirow}
\usepackage{xcolor}

\usepackage{mathtools,tikz,caption}
\DeclareRobustCommand\sampleline[1]{\tikz\draw[#1] (0,0) (0,\the\dimexpr\fontdimen22\textfont2\relax)
  -- (1em,\the\dimexpr\fontdimen22\textfont2\relax);}

\title{SCD: Self-Contrastive Decorrelation for Sentence Embeddings}

\author{Tassilo Klein \\
  SAP AI Research \\
  \texttt{tassilo.klein@sap.com} \\\And
  Moin Nabi \\
    SAP AI Research \\
  \texttt{m.nabi@sap.com} \\}

\begin{document}
\maketitle
\begin{abstract}
In this paper, we propose Self-Contrastive Decorrelation (\ours), a self-supervised approach. Given an input sentence, it optimizes a joint self-contrastive and decorrelation objective. Learning a representation is facilitated by leveraging the contrast arising from the instantiation of standard dropout at different rates. The proposed method is conceptually simple yet empirically powerful. It achieves comparable results with state-of-the-art methods on multiple benchmarks without using contrastive pairs. This study opens up avenues for efficient self-supervised learning methods that are more robust than current contrastive methods.\footnote{Source code and pre-trained models are available at: \url{https://github.com/SAP-samples/acl2022-self-contrastive-decorrelation/}}

\end{abstract}

\section{Introduction}
Unsupervised learning of representation (a.k.a. embedding) is a fundamental problem in NLP and has been studied extensively in the literature~\citep{mikolov2013efficient, pennington-etal-2014-glove, mccann2017learned, peters2018deep}. Sentence embeddings are essential for numerous language processing applications, such as machine translation, sentiment analysis, information retrieval, and semantic search.
Recently, self-supervised pre-training schemes have been successfully used in the context of transformer architectures, leading to a paradigm shift in natural language processing and understanding~\citep{devlin2018bert,liu2019roberta,radford2018improving}
The idea here is to employ an auxiliary task, which enforces an additional objective during training. Typically, this entails predictions based on a subset of information from the context.
Most objectives found effective in practice are quite simple.
Some successful examples of such pretext tasks are Masked Language Model (MLM), Next Sentence Prediction (NSP), Sentence Order Prediction (SOP), etc.~\citep{devlin-etal-2019-bert,liu2019roberta,lan2019albert}.
When working with unlabeled data, contrastive learning is among the most powerful approaches in self-supervised learning. 
The goal of contrastive representation learning is to learn an embedding space in such a manner that similar sample pairs (i.e., \emph{positive pairs}) stay close to each other. Simultaneously, dissimilar sample pairs (i.e., \emph{negative pairs}) are far pushed apart. To this end, different augmented views of the same sample and the augmented views from different samples are used as positive and negative pairs. These methods have shown impressive results over a wide variety of tasks from visual to textual representation learning~\citep{chen2020simple, chen2020improved, gao2021simcse, grill2020bootstrap,chen2021exploring}.

Different techniques have been proposed for the augmentation and selection of positive and negative pairs. For example,  DeCLUTR~\cite{giorgi2020declutr} proposes to take different spans from the same document as positive pairs, while CT~\cite{carlsson2020semantic} aligns embeddings of the same sentence from two different encoders.
CERT~\cite{fang2020cert} applies the back-translation to create augmentations of original sentences, and IS-BERT~\cite{zhang-etal-2020-unsupervised} maximizes the agreement between global and local features. Finally, CLEAR~\cite{wu2020clear} employs multiple sentence-level augmentation strategies to learn a sentence representation. Despite the simplicity of these methods, they require careful treatment of negative pairs, relying on large batch sizes \citep{chen2020simple} or sophisticated memory strategies. These include memory
banks~\cite{chen2020improved,He_2020_CVPR} or customized mining strategies \cite{klein2020contrastive} to retrieve negative pairs efficiently. 
In NLP specifically, the endeavor of ``hard negative mining'' becomes particularly challenging in the unsupervised scenario. Increasing training batch size or the memory bank size implicitly introduces more hard negative samples, coming along with the heavy burden of large memory requirements. 

In this paper, we introduce \ours, a novel algorithm for self-supervised learning of sentence embedding. 
\ours achieves comparable performance in terms of sentence similarity-based tasks compared with state-of-the-art contrastive methods \emph{without}, e.g., employing explicit contrastive pairs. Rather, in order to learn sentence representations, the proposed approach leverages the \emph{self}-contrast imposed on the augmentations of a \emph{single} sample. In this regard, the approach builds upon the idea that sufficiently strong perturbation of the sentence embedding reflects the semantic variations of the sentence. However, it is unclear which perturbation is simply a slight variation of the sentence without changing the semantic (positive pair) and which perturbation sufficiently modifies the semantic to create a negative sample. Such ambiguity manifests itself in the augmented sample sharing the characteristics of both negative and positive samples. To accommodate this, we propose an objective function consisting of two opposing terms, which acts on augmentations pairs of a sample:  \textbf{i)} self-contrastive divergence (\emph{repulsion}), and \textbf{ii)} feature decorrelation (\emph{attraction}). 
The first term treats the two augmentations as a negative pair pushing apart the different views. In contrast to that, the second term attends to the augmentations as a positive pair. Thus, it maximizes the correlation of the same feature across the views, learning invariance w.r.t. the augmentation. Given the opposing nature of the objectives, integrating them in a joint loss yields a min-max optimization scheme.
The proposed approach avoids degenerated embeddings by framing the representation learning objective as an attraction-repulsion trade-off. Simultaneously, it learns to improve the semantic expressiveness of the representation.
Due to the difficulty of augmentation in NLP, the proposed approach generates augmentation ``on-the-fly'' for each sample in the batch. To this end, multiple augmentations are produced by \emph{varying} dropout rates for each sample. We empirically observed that \ours is more robust to the choice of augmentations than pairwise contrastive methods; we
believe that not relying on contrastive pairs is one of the main reasons for this, an observation also made in self-supervised learning literature such as BYOL~\cite{grill2020bootstrap}.
While other methods take different augmentation or different copies of models, we utilized the different outputs of the same sentence from standard dropout. 

Most related to our paper is \citep{gao2021simcse}, which considers using dropout as data augmentation in the context of contrastive learning. A key novelty of our approach is that we use the dropout for creating the \emph{self-}contrastive pairs, which can be utilized as \emph{both} positive and negative.
At last, we note that our model is different from the \emph{pairwise} feature decorrelation or whitening in \citep{zbontar2021barlow, su2021whitening, ermolov2021whitening}, which
encourage similar representations between augmented views of a sample while minimizing the redundancy within the representation vector. A key difference compared to these methods
is that they ignore the contrastive objective completely. In contrast, our method takes it into account and provides the means to treat self-contrastive views as positive and negative pairs simultaneously. 

\vspace{0.5mm}
\textbf{Our contribution:} \textbf{i)} generation of sentence embeddings by leverage multi-dropout \textbf{ii)} elimination of reliance on negative pairs using self-contrast, \textbf{iii)} proposing feature decorrelation objective for non-contrastive self-supervised learning in NLP.

\begin{figure*}[h!]
    \centering
    \includegraphics[width=1.0\textwidth]{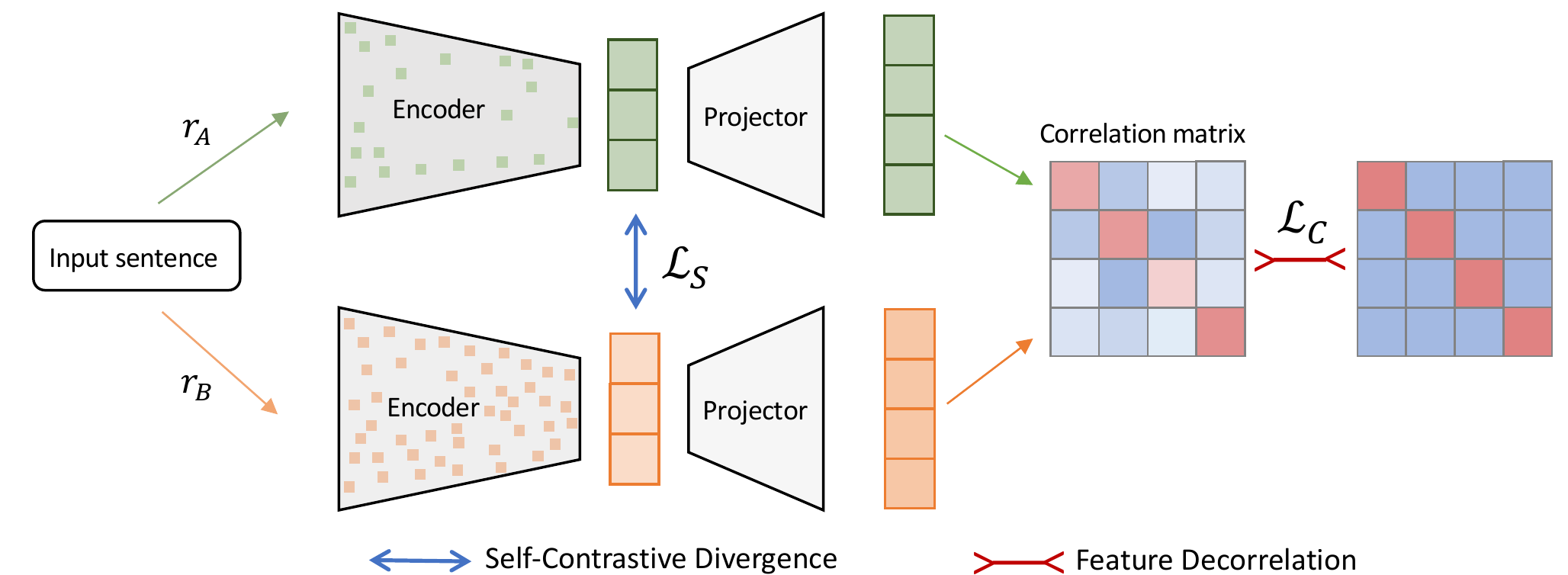}
    \caption{Schematic illustration of the proposed approach (best shown in color). Starting from an input sentence \textbf{(left)}, two embeddings are produced by varying the dropout-rate in the encoder. Patches within the encoder indicate masking due to dropout. Different dropout rates and resulting embeddings color-coded: \textcolor{ForestGreen}{\textbf{low dropout}}, \textcolor{Bittersweet}{\textbf{high dropout}}. Self-contrastive loss is imposed on the embeddings \textbf{(center)}. A projector maps embeddings to to a high-dimensional feature space, where the features are decorrelated \textbf{(right)}.}
    \label{fig:method}
\end{figure*}

\vspace{-0.5mm}
\section{Method}
Our approach relies on the generation of two views $A$ and $B$ of samples. To this end, augmentations are generated in embedding space for each sample $x_i$ in batch $X$. Batches are created from samples of set $\mathcal{D}=\{(x_i)\}_{i=1}^N$, where $N$ denotes the number of sample (sentences). Augmentations are produced by an encoder $f_\theta$, parametrized by $\theta$. The output of the encoder is the embeddings of samples in $X$ denoted as $H^A\in \mathcal{T}$ and $H^B\in \mathcal{T}$. Here $\mathcal{T}$ denotes the embedding space. Next, we let, $\vh_i \in \mathcal{T}$ denote the associated representation of the sentence. The augmentation embeddings produced per sample are then denoted $\vh^A_i$ and $\vh^B_i$. To obtain the different embedding, we leverage a transformer language model as an encoder in combination with \emph{varying} dropout rates. Specifically, one augmentation is generated with \emph{high} dropout and one with \emph{low} dropout. This entails employing different random masks during the encoding phase. The random masks are associated with \emph{different} ratios, $r_A$ and $r_B$, with $r_A < r_B$. Integrating the distinct dropout rates into the encoder, we yield $\vh^A_i=f_\theta(x_i, r_A)$ and $\vh^B_i=f_\theta(x_i, r_B)$.
Given the embeddings, we leverage a joint loss, consisting of two objectives:
\begin{equation}
\label{eq:NZG}
\begin{aligned}
\min_{\theta_1, \theta_2} \mathcal{L}_{S}(f_{\theta_1})+ \alpha\mathcal{L}_{C}(f_{\theta_1},p_{\theta_2})
\end{aligned}
\end{equation}
Here $\alpha \in \mathbb{R}$ denotes a hyperparameter and $p: \mathcal{T} \to \mathcal{P}$ is a projector (MLP) parameterized by $\theta_2$, which maps the embedding to $\mathcal{P}$, with $|\mathcal{P}| \gg |\mathcal{T}|$.
\\
The objective of $\mathcal{L}_{S}$ is to increase the contrast of the augmented embedding, pushing apart the embeddings $\vh^A_i$ and $\vh^B_i$.
The objective of $\mathcal{L}_{C}$ is to reduce the redundancy and promote invariance w.r.t. augmentation in a high-dimensional space $\mathcal{P}$. See Fig.~\ref{fig:method} for a schematic illustration of the method.

\vspace{-1mm}
\subsection{Self-Contrastive Divergence:}
Self-contrast seeks to create a contrast between the embeddings arising from different dropouts. Hence, $ \mathcal{L}_{S}$ consists of the cosine similarity of the samples in the batch as:
\begin{equation}
    \mathcal{L}_{S} = \frac{1}{N}\sum^N_i {\vh_i^A \cdot (\vh_i^B)^T}{\left(\|\vh_i^A\|  \|\vh_i^B\|\right)^{-1}}
\end{equation}

\subsection{Feature Decorrelation:}
$\mathcal{L}_{C}$ seeks to make the embeddings invariant to augmentation while at the same time reducing the redundancy in feature representation. To this end, the embedding $\vh_i$ is projected up from $\mathcal{T}$ to a high-dimensional space $\mathcal{P}$, where decorrelation is performed. To avoid clutter in notation, we let $p^*_i = p(h^*_i)$ and $* \in \{A, B\}$, denote the augmented embedding vectors of sample $x_i$ after applying a projection with $p(.)$. Then, a correlation matrix is computed from the projected embeddings. Its entries $C_{j,k}$ are:
\begin{equation}
    C_{jk} = {\sum_i p^A_{i,j} \cdot p^B_{i,k}}\left({\sum_i (p^A_{i,j})^2 (p^B_{i,k})^2}\right)^{-\frac{1}{2}}
\end{equation}
Here, $p^*_{i,j} \in \mathbb{R}$ denotes the $j^{th}$ component in the projected embedding vector. Then the loss objective for feature decorrelation is defined as:
\begin{equation}
     \mathcal{L}_{C} = -\sum_j (1- C_{jj})^2 + \lambda \sum_j\sum_{j\neq k} C_{jk}^2 
\end{equation}
The first term seeks to achieve augmentation invariance by maximization of the cross-correlation along the diagonal. The second term seeks to reduce redundancy in feature representation by minimizing correlation beyond the diagonal.
Given that these objectives are opposing, $\lambda \in \mathbb{R}$ is a hyperparameter, controlling the trade-off.

\begin{table*}[t!]
    \begin{center}
    \centering
    \small
\begin{tabular}{lcccccccc}
    \toprule

        \multicolumn{9}{c}{\it{Semantic Textual Similarity (STS) Benchmark}}\\
         \midrule
         \tf{Model} & \tf{STS12} & \tf{STS13} & \tf{STS14} & \tf{STS15} & \tf{STS16} & \tf{STS-B} & \tf{SICK-R} & \tf{Avg.} \\
   
    \midrule
         GloVe embeddings (avg.)$^\clubsuit$ & 55.14 & 70.66 & 59.73 & 68.25 & 63.66 & 58.02 & 53.76 & 61.32 \\
         BERT\ba \cls-embedding & 21.54&	32.11&	21.28&	37.89&	44.24&	20.29&	42.42&	31.40\\ BERT\ba (first-last avg)$^\lozenge$ & 39.70 & 59.38 & 49.67 & 66.03 & 66.19 & 53.87 & 62.06 & 56.70 \\
         BERT\ba-flow$^\lozenge$ & 58.40&	67.10&	60.85&	75.16&	71.22&	68.66&	64.47&	66.55 \\ BERT\ba-whitening$^\lozenge$ & 57.83& 66.90 & 60.90 & 75.08& 71.31& 68.24& 63.73& 66.28\\ IS-BERT\ba$^\heartsuit$ & 56.77 & 69.24 & 61.21 & 75.23 & 70.16 & 69.21 & 64.25 & 66.58 \\
         CT-BERT\ba$^\lozenge$ & 61.63 & 76.80 & 68.47 & 77.50 & 76.48 & 74.31 & 69.19 &72.05 \\
SimCSE-BERT\ba & \tf{68.05}&	\tf{80.38} &	\tf{72.62}&	\tf{78.96}&	\tf{76.90}&	{75.11}&	{69.37}&	\tf{74.48} \\
        $*$ \ours-BERT\ba & 66.94 & 78.03 & 69.89 & 78.73 & 76.23 & \tf{76.30} & \tf{73.18} & 74.19 \\
       \hdashline RoBERTa\ba\cls-embedding & 16.67&	45.56&	30.36&	55.08&	56.99&	38.82&	61.89&	43.62\\ RoBERTa\ba (first-last avg)$^\lozenge$ & 40.88 & 58.74 & 49.07 & 65.63 & 61.48 & 58.55 & 61.63 & 56.57 \\
         SimCSE-RoBERTa\ba & \tf{67.05}&	\tf{80.01} &	\tf{70.93}&	79.66&	\tf{80.06}&	\tf{78.38}&	{68.30}&	\tf{74.91} \\
          $*$ \ours-RoBERTa\ba & 63.53 & 77.79 & 69.79 & \tf{80.21} & 77.29 & {76.55} & \tf{72.10} & 73.89 \\
    \bottomrule
    \end{tabular}
\end{center}

    \caption{
        Sentence embedding performance on STS tasks measured as Spearman’s correlation.
        $\clubsuit$: results from \citet{reimers-gurevych-2019-sentence};
        $\heartsuit$: results from \citet{zhang-etal-2020-unsupervised};
        $\lozenge$ results from \citet{gao2021simcse};
        other results are by ourselves. Dashed line (\sampleline{dashed}), separates BERT (upper part) and RoBERTa (lower part) language models.
    }
    \label{tab:main_sts}
    \vspace{-5pt}
\end{table*}

\section{Experiments \& Results}
\subsection{Training Setup:}
Training is started from a pre-trained transformer LM. Specifically, we employ the Hugging Face~\cite{Wolf2019HuggingFacesTS} implementation of BERT and RoBERTa. 
For sentence representation, we take the embedding of the {\tt[CLS]} token. Then similar to~\cite{gao2021simcse}, we train the model in an unsupervised fashion on $10^6$ randomly samples sentences from Wikipedia. The LM is trained with a learning rate of $3.0\mathrm{e}{-5}$ for $1$ epoch at batch-size of $192$. The projector MLP $q$ has three linear layers, each with 4096 output units in conjunction with ReLU and BatchNorm in between.
For BERT hyperparameters are $\alpha=0.005$, $\lambda=0.013$, and dropout rates are $r_A=5.0\%$ and $r_B=15.0\%$.  For RoBERTa hyperparameters are $\alpha=0.0033$, $\lambda=0.028$, and dropout rates are $r_A=6.5\%$ and $r_B=24.0\%$. 
The values were obtained by grid-search. First a coarse-grid was put in place with a step-size  of $0.1$
for $\alpha$, $10\%$ for the dropout rates $r_A, r_B$. For $\lambda$ the coarse-grid consisted of different magnitudes $\{0.1,0.01,0.001\}$. Second, on a fine-grid with step-size of $0.01$ and $1\%$, respectively. 

\subsection{Evaluation Setup:}
Experiments are conducted on 7 standard semantic textual similarity (STS) tasks. In addition to that, we also evaluate on 7 transfer tasks. Specifically, we employ the SentEval toolkit~\cite{conneau-kiela-2018-senteval} for evaluation. As proposed by ~\cite{reimers-gurevych-2019-sentence,gao2021simcse}, we take STS results as the main
comparison of sentence embedding methods and transfer task results for reference. For the sake of comparability, we follow the evaluation protocol of~\cite{gao2021simcse}, employing Spearman’s rank correlation and aggregation on all the topic subsets.

\begin{table*}[t]
    \begin{center}
    \centering
    \small
\begin{tabular}{lcccccccc}
    \toprule

        \multicolumn{9}{c}{\it{Transfer Benchmark}}\\
    \midrule
        \tf{Model} & \tf{MR} & \tf{CR} & \tf{SUBJ} & \tf{MPQA} & \tf{SST} & \tf{TREC} & \tf{MRPC} & \tf{Avg.}\\
    \midrule
         GloVe embeddings (avg.)$^\clubsuit$ & 77.25&    78.30&  91.17&  87.85&  80.18&  83.00& 72.87 & 81.52\\
         Skip-thought$^\heartsuit$ &  76.50& 80.10&  93.60&  87.10&  82.00&  92.20&  73.00& 83.50  \\
        \midrule
         Avg. BERT embeddings$^\clubsuit$ & 78.66 & 86.25 & 94.37 & 88.66 & 84.40 & \tf{92.80} & 69.54 & 84.94 \\
         BERT \cls-embedding &  \tf{81.83} &  \tf{87.39} & 95.48 & 88.21 & \tf{86.49} & 91.00 & 72.29 & \tf{86.10} \\
         IS-BERT\ba$^\heartsuit$ & 81.09 & 87.18 & 94.96 & \tf{88.75} & 85.96 & 88.64 & 74.24 & 85.83 \\
SimCSE-BERT\ba & 80.74&	85.75&	93.96&	88.60&	84.57&	86.20&	73.51&	84.76\\
         $*$ \ours-BERT\ba & 73.21 & 85.80 & \tf{99.56} & 88.67 & 85.89 & 89.80 & \tf{75.71} & 85.52 \\ 
          \hdashline
          RoBERTa \cls-embedding & 81.27 & 84.77 & 94.15 & 84.18 & 86.71 & 81.20 & 72.17 & 83.49 \\
          SimCSE-RoBERTa\ba & 65.00&	87.28&	\tf{99.60}&	\tf{86.63}&	87.26&	80.80&	72.23&	82.69\\
         $*$ \ours-RoBERTa\ba & \tf{82.17} & \tf{87.76} & {93.67} & 85.69 & \tf{88.19} & \tf{83.40} & \tf{76.23} & \tf{85.30} \\ 
    \bottomrule
    \end{tabular}
\end{center}

    \caption{
        Transfer task result measured as accuracy.
        $\clubsuit$: results from \citet{reimers-gurevych-2019-sentence};
        $\heartsuit$: results from \citet{zhang-etal-2020-unsupervised};
        $\lozenge$ results from \citet{gao2021simcse};
        other results are by ourselves. Dashed line (\sampleline{dashed}), separates BERT (upper part) and RoBERTa (lower part) language models.
    }
    \label{tab:transfer_sts}
    \vspace{-5pt}
\end{table*}

 \begin{figure}[b!]
    \centering
    \includegraphics[width=0.495\textwidth]{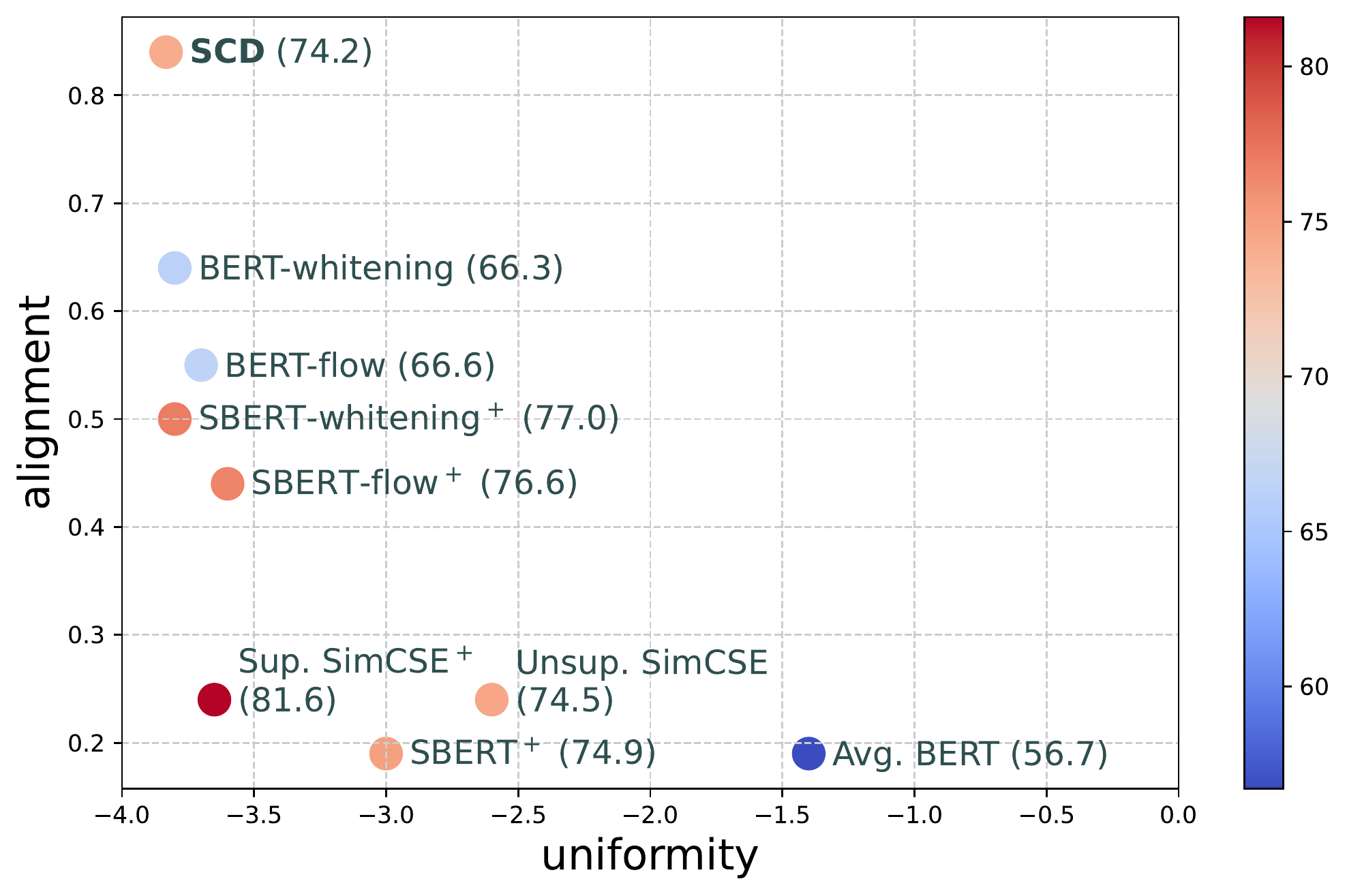}
    \caption{Quantitative analysis of embeddings - \emph{alignment} vs. \emph{uniformity} (the smaller, the better). Points represent average STS performance using BERT\ba, with Spearman’s correlation color coded ($+$ corresponds to supervised methods).} 
\label{fig:alignment_vs_uniformity}
\end{figure}

\subsection{Main Results}
\subsubsection{Semantic Textual Similarity:} 
We evaluate on 7 STS tasks: ~\cite{agirre-etal-2012-semeval,agirre-etal-2013-sem,agirre-etal-2014-semeval,agirre-etal-2015-semeval,agirre-etal-2016-semeval}, 
STS Benchmark~\cite{cer-etal-2017-semeval}  and
SICK-Relatedness~\cite{marelli-etal-2014-sick}. These datasets come in sentence pairs together with correlation labels in the range of 0 and 5, indicating the semantic relatedness
of the pairs. Results for the sentence similarity experiment can be seen in Tab.~\ref{tab:main_sts}. The proposed approach is on-par with state-of-the-art approaches. Using BERT-LM, we outperform the next-best approach on STS-B $(\textbf{+1.19})$ and on SICK-R $(\textbf{+3.81})$ points. Using RoBERTa-LM, we outperform  the next best comparable approach (SimCSE-RoBERTA\ba) on STS-15 $(\textbf{+0.55\%})$ and SICK-R $(\textbf{+3.8\%})$.

\subsubsection{Transfer task:}
We evaluate our models on the following transfer tasks: MR~\cite{pang2005seeing_mr}, CR~\cite{hu2004mining_cr}, SUBJ~\cite{pang2004sentimental_subj}, MPQA~\cite{wiebe2005annotating_mpqa}, SST-2~\cite{socher2013recursive_sst-2}, TREC~\cite{voorhees2000building_trec} and MRPC~\cite{dolan-brockett-2005-automatically-mrpc}. To this end, a logistic regression classifier is trained on top of (frozen) sentence embeddings produced by different methods. We follow default configurations from SentEval. Results for the transfer task experiment can be seen in Tab.~\ref{tab:transfer_sts}.
\ours is on-par with state-of-the-art approaches.
Using BERT-LM, we outperform the next best approach on SUBJ $(\textbf{+4.6\%})$ and MRPC $(\textbf{+2.2\%})$. Using RoBERTa-LM, we outperform the next best comparable approach (SimCSE-RoBERTA\ba) on almost all benchmarks, with an average margin of $(\textbf{+2.61\%})$.

\subsection{Analysis}

\subsubsection{Ablation Study:}
We evaluated each component's performance by removing them individually from the loss to assess both loss terms' contributions. It should be noted that $\mathcal{L}_{S}$ of Eq. 2 and $\mathcal{L}_{C}$ of Eq. 4 both interact in a competitive fashion. Hence, only the equilibrium of these terms yields an optimal solution. Changes - such as eliminating a term - have detrimental effects, as they prevent achieving such an equilibrium, resulting in a significant drop in performance. See Tab. \ref{tab:ablation-study} for the ablation study on multiple benchmarks. Best performance is achieved in the presence of all loss terms.
\\
\begin{table}[b!]
\centering
  \small
\begin{tabular}{l|l l l}
 \toprule
\textbf{Method} & \textbf{STS} & \textbf{STS-B} & \textbf{SICK-R}\\
 \midrule
\centering
BERT-\ba & 31.41 & 20.29 & 42.42\\
\ours (\texttt{$\mathcal{L}_{S}$}) & 35.70 & 23.59 & 49.88\\
\ours (\texttt{$\mathcal{L}_{C}$}) & 66.48 & 67.57 & 67.97\\
\midrule
\ours (\texttt{$\mathcal{L}_{S}$+$\mathcal{L}_{C}$}) & \textbf{73.96} & \textbf{76.30} & \textbf{73.18}\\
  \bottomrule
\end{tabular}
\caption{Ablation study, performance in average Spearman correlation on Semantic Texual Similarity task. \textbf{STS} denotes the average of STS12 to STS16.}
\label{tab:ablation-study}
\end{table}
\subsubsection{Uniformity and Alignment Analysis:}
To better understand the strong performance of \ours, we borrow the analysis tool from \citep{wang2020understanding}, which takes \emph{alignment} between semantically-related positive pairs and \emph{uniformity} of the whole representation space to measure the quality of learned embeddings. Figure \ref{fig:alignment_vs_uniformity} shows \emph{uniformity} and \emph{alignment} of different methods and their results on the STS. \ours achieves the best in terms of \emph{uniformity}, reaching to the supervised counterparts (\textbf{-3.83}), which can be related to the strong effect of the self-contrastive divergence objective. It shows the \emph{self}-contrastive pairs can effectively compensate for the absence of contrastive pairs. In terms of \emph{alignment}, \ours is inferior to other counterparts (\textbf{0.84}), which can be attributed to the fact that our repulsion objective mainly focuses on the feature decorrelation aiming to learn a 
more effective and efficient representation. This is reflected in the final results on the STS where \ours obtains significantly higher correlation even compared to the method with lower \emph{alignment} such as BERT-whitening or BERT-flow.

\section{Conclusion \& Future Work}
We proposed a self-supervised representation learning approach, which leverages the self-contrast of augmented samples obtained by dropout.  Despite its simplicity, it achieves comparable results with state-of-the-arts on multiple benchmarks. Future work will deal with sample-specific augmentation to improve the embeddings and, particularly, the representation alignment.
\\

\noindent\textbf{Acknowledgement: }
We would like to thank Mahdyar Ravanbakhsh for valuable feedback on the manuscript.
\bibliography{anthology,custom}
\bibliographystyle{acl_natbib}

\end{document}